\documentclass{article}
\usepackage[position,final]{neurips_2026}

\usepackage{tikz}
\usetikzlibrary{arrows.meta,positioning,shapes.geometric,fit,backgrounds}
\usepackage{enumitem}
\usepackage{algorithm}
\usepackage{algorithmic}
\usepackage[utf8]{inputenc} 
\usepackage[T1]{fontenc}    
\usepackage{hyperref}       
\usepackage{url}            
\usepackage{booktabs}       
\usepackage{amsmath, amssymb, amsthm}     
\usepackage{nicefrac}       
\usepackage{microtype}      
\usepackage{xcolor}         

\usepackage{wrapfig}
\usepackage{soul,bm}
\usepackage{dsfont}
\newtheorem{theorem}{Theorem}[section]
\newtheorem{proposition}{Proposition}

\theoremstyle{definition}

\newtheorem{assumption}{Assumption}

\theoremstyle{remark}
\newtheorem{remark}[theorem]{Remark}

\newcommand{\coloneqq}{\mathrel{:=}}

\title{The Post-Training Dilemma: Why We Should Rethink the Sequential SFT-RL Paradigm}

\author{%
  Xueyan Niu  \And Bo Bai  \And Wei Han \And Weixi Zhang \\
\AND
  \vspace{-2em}
  \phantom{XXX}\\
  Theory Lab, 2012 Labs, Huawei Technologies Co., Ltd. \\
  \texttt{\{niuxueyan3, bobai8, harvey.hanwei, zhangweixi1\}@huawei.com} \\
}

\begin{document}

\maketitle

\begin{abstract}

This position paper argues that the sequential supervised fine-tuning followed by reinforcement learning (SFT-then-RL) pipeline---the de facto paradigm for large language model (LLM) post-training---has a fundamental property that cannot be fixed by hyperparameter tuning: SFT and RL optimize over nearly orthogonal parameter subspaces, so RL inevitably degrades the capabilities acquired during the preceding SFT phase. Under the Polyak-Łojasiewicz (PL) condition, which is empirically observed for over-parameterized networks, we show that RL increases the SFT loss by at least $\frac{\mu}{2}\mathrm{dist}(\theta_{\mathrm{RL}}, \Theta^*_{\mathrm{SFT}})^2$ for some $\mu>0$. 
Although the KL penalty can constrain distributional shift, it cannot correct the underlying gradient misalignment between the two objectives.
Experiments on Qwen3-0.6B across four tasks confirm the theoretical predictions: on MATH-500, Alpaca, HH-RLHF, and SST-2, the SFT loss increases after the RL phase. 
We identify the underlying mechanism as persistent near-orthogonality between the gradients of the two objectives, paired with opposing landscape dynamics: SFT flattens the loss
landscape while RL re-sharpens it.
Based on our analysis, we propose three promising research directions: (1) joint optimization of the two objectives with gradient surgery to resolve conflicting updates; (2) landscape-aware scheduling that terminates RL training at the theoretically optimal point; and (3) characterizing the model scale threshold above which SFT becomes unnecessary.

\end{abstract}

\section{Introduction}

Nearly all large language models (LLMs) deployed currently rely on a two-stage post-training pipeline: supervised fine-tuning (SFT) followed by reinforcement learning (RL) \citep{christiano2017deep,ouyang2022training,touvron2023llama,deepseekr1}. The implicit assumption underlying this paradigm is that these stages are functionally composed: SFT teaches basic instruction-following capabilities, RL then refines the model to align with human preferences~\cite{ouyang2022training} or optimize verifiable reward signals~\cite{shao2024grpo}, and the final model deployed inherits the capabilities of both stages (See Figure~\ref{fig:pipeline}).

This pervasive assumption does not always hold. Under the Polyak-Łojasiewicz (PL) condition~\citep{polyak1963gradient,karimi2016}, a condition weaker than strong convexity that is known to hold for over-parameterized neural networks~\citep{zhang2016understanding,aich2025lplr}, we prove that RL inevitably degrades the performance achieved during the preceding SFT phase:
$\mathcal{L}_\mathrm{SFT}(\theta_\mathrm{RL}) \geq \mathcal{L}_\mathrm{SFT}(\theta_\mathrm{SFT}^*) + 
\frac{\mu_{\mathrm{SFT}}}{2}\,
\mathrm{dist}(\theta_{\mathrm{RL}}, \Theta^*_{\mathrm{SFT}})^2$.
The mechanism driving this degradation is gradient misalignment: the gradients of the SFT and RL objectives are nearly orthogonal ($\cos(g_\mathrm{SFT}, g_\mathrm{RL}) \approx 0$), suggesting that the two stages optimize over largely disjoint parameter subspaces. 
This is paired with opposing landscape dynamics: SFT reduces the SFT gradient norm by an order of magnitude, while RL re-sharpens the landscape by increasing the gradient norm.
While the KL penalty can constrain the magnitude of parameter displacement, it cannot correct the fundamental directional misalignment between the two objectives.

DeepSeek-R1~\citep{deepseekr1} bypasses coupling by eliminating SFT at 671B MoE scale. 
Concurrent work~\citep{chen2025sftorrl,chen2026synergydilemma,sft-memorizes} validates similar findings from complementary angles.
The more recent post-training recipe \citep{wang2025nemotron} applies cascaded, domain-wise RL on top of a broad multi-stage SFT.
Empirical evidence such as \cite{liu2025acereason} suggests that a stronger SFT model consistently leads to better final performance after large-scale RL training.
Notably, models often experience ``catastrophic forgetting'' during the transition from SFT to RL \cite{chen2025beyond}.
Previous work \cite{liu2025acereason} applied RL to a series of different SFT models and found that RL unlocks new abilities that the starting SFT models do not possess.
However, \cite{yan2025learning} showed that on-policy RL amplifies existing behaviors instead of introducing new capacities. 
In \cite{chen2026synergydilemma}, the authors proposed the ``synergy dilemma'' for large vision-language models by comparing three post-training strategies: two-stage SFT \& RL, interleaved SFT \& RL, and progressive SFT \& RL. Their results show no synergy between SFT and RL in multi-modal reasoning. 
In \cite{he2025justrl}, a single RL stage is used to achieve state-of-the-art performance on two distilled 1.5B reasoning models.
However, no existing method addresses the gradient misalignment at its root. The field needs either coupling-aware joint optimization or a principled understanding of when single-phase training suffices.

\textbf{Our position} is that the sequential SFT-then-RL pipeline needs a paradigmatic shift. The coupling phenomenon is structural, and no choice of the KL penalty coefficient $\beta$, RL algorithm, or training duration can eliminate this inherent degradation. We argue that the field must move toward three viable alternative paradigms:
(1) joint optimization of the two objectives with gradient surgery to remove conflicting update components;
(2) landscape-aware adaptive scheduling that stops RL training before cumulative SFT damage overwhelms the alignment benefits; and
(3) characterizing the model scale threshold above which single-phase pure RL training becomes sufficient.

\textbf{Our contributions} are as follows.
\begin{itemize}
\item We derive PL-condition lower bounds that formalize the non-decoupling phenomenon (Theorem~\ref{thm:sft-rl}, Propositions~\ref{prop:optimal-steps}--\ref{prop:threshold}), proving that the coupling between SFT and RL is an inherent structural property of the sequential pipeline.
\item We conduct experiments across diverse tasks, observing SFT loss increases after RL. We identify the underlying mechanism as 
persistent gradient near-orthogonality and a measurable seesaw
dynamic in the SFT landscape (flattened by SFT, re-sharpened by RL).
\item We outline a concrete research roadmap to move beyond the sequential paradigm, including joint optimization with gradient surgery, landscape-aware adaptive scheduling, and characterizing the SFT-necessity scale threshold.
\end{itemize}

\begin{figure}[t]
\centering
\includegraphics[width=\textwidth]{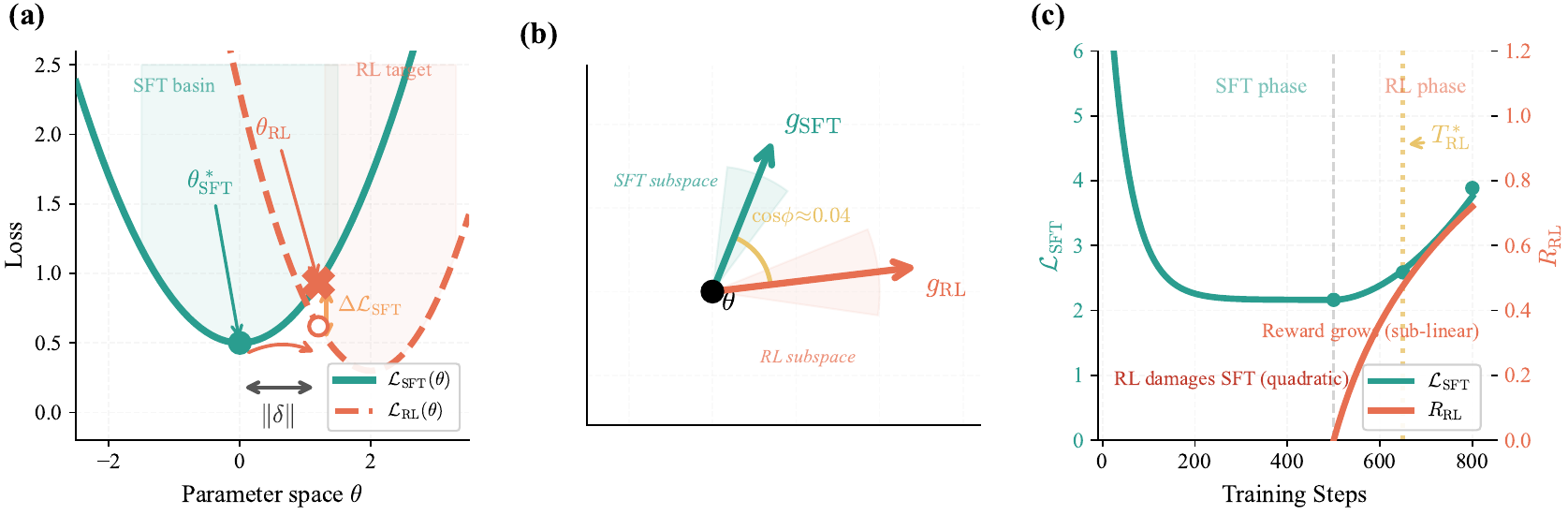}
\caption{\textbf{Mechanism of non-decoupling in sequential SFT$\to$RL post-training.} \textbf{(a)~Parameter-space view:} RL displaces parameters from the SFT minimizer $\theta_{\mathrm{SFT}}^\ast\in \Theta_{\mathrm{SFT}}^\ast$ by the RL-induced displacement $\|\delta\|$, causing quadratic SFT loss increase $\Delta\mathcal{L}_{\mathrm{SFT}} \geq \frac{\mu}{2}\|\delta\|^2$ (Theorem~\ref{thm:sft-rl}). The SFT and RL loss landscapes have minima in different regions, so RL moves parameters away from $\theta_{\mathrm{SFT}}^\ast$. \textbf{(b)~Gradient-space view:} conceptual misalignment between the SFT and RL gradient directions. In practice, the two directions are nearly orthogonal (see Figure~\ref{fig:gradient-dynamics} for the measured gradient dynamics), meaning RL updates displace parameters in a direction largely irrelevant to SFT preservation. \textbf{(c)~Training dynamics:} SFT loss ($\mathcal{L}_{\mathrm{SFT}}$, left axis) decreases during the 500-step SFT phase, then increases during the 300-step RL phase (COLA, Qwen3-0.6B). Meanwhile, RL reward ($R_{\mathrm{RL}}$, right axis) grows only sub-linearly. The optimal stopping point $T_{\mathrm{RL}}^\ast$ (Proposition~\ref{prop:optimal-steps}) balances RL benefit against SFT damage.}
\label{fig:non-decoupling-mechanism}
\end{figure}

Figure~\ref{fig:non-decoupling-mechanism} illustrates the three perspectives underlying non-decoupling: parameter-space displacement, gradient-space misalignment, and the resulting training dynamics where SFT loss inevitably rises during the RL phase.
We explicitly address potential alternative views to our thesis in Section~\ref{sec:objections}, and we state clear falsifiable predictions that allow the community to test our theory. 
If non-decoupling proves absent at scale, this would be an important negative result that constrains the theory's scope. If it persists, the community faces a question worth debating: \textit{to what extent should we accept unavoidable capability degradation as the cost of alignment, and when does that cost become unacceptable?}

\begin{figure}[t]
\centering
\begin{tikzpicture}[
    >=Stealth,
    node distance=1.0cm and 1.2cm,
    block/.style={rectangle, draw, rounded corners, minimum height=0.85cm, minimum width=2.0cm, align=center, font=\small},
    phase/.style={block, fill=blue!10, minimum width=2.3cm},
    result/.style={block, fill=orange!10},
    insight/.style={rectangle, draw, dashed, rounded corners, fill=green!8, minimum height=0.7cm, minimum width=1.8cm, align=center, font=\tiny},
]
\node[phase] (pre) {Pre-trained\\$\theta_0$};
\node[phase, right=of pre] (sft) {SFT Phase\\$\theta_0 \to \theta_\mathrm{SFT}$};
\node[phase, right=of sft] (rl) {RL Phase\\$\theta_\mathrm{SFT} \to \theta_\mathrm{RL}$};
\node[result, right=of rl] (final) {Final Model\\$\theta^*$};

\draw[->,thick] (pre) -- (sft);
\draw[->,thick] (sft) -- (rl);
\draw[->,thick] (rl) -- (final);

\draw[->,thick,red!70!black,bend left=35] (rl.south) to node[above,font=\tiny,red!70!black,text width=2.3cm,align=center] {non-decoupling:\\$\Delta\mathcal{L}_\mathrm{SFT} \geq \frac{\mu}{2}\|\delta\|^2$} (sft.south);

\node[insight, below=0.65cm of sft] (i2) {$\cos(g_\mathrm{SFT},g_\mathrm{RL})\!\approx\!0$\\gradient misalignment};
\node[insight, below=0.65cm of rl] (i3) {SFT loss increase\\(Theorem~\ref{thm:sft-rl})};
\node[insight, below=0.65cm of final] (i4) {Optimal $T_\mathrm{RL}^*$\\(Proposition~\ref{prop:optimal-steps})};

\draw[->,gray,thin] (i2) -- (sft);
\draw[->,gray,thin] (i3) -- (rl);
\draw[->,gray,thin] (i4) -- (final);
\end{tikzpicture}
\caption{The sequential SFT$\to$RL post-training pipeline and the non-decoupling phenomenon. The red curved arrow shows the argument: RL optimization degrades SFT performance ($\Delta\mathcal{L}_\mathrm{SFT} \geq \frac{\mu}{2}\|\delta\|^2$, Theorem~\ref{thm:sft-rl}). 
}
\label{fig:pipeline}
\end{figure}

\section{The Non-Decoupling Phenomenon}

\subsection{Preliminary}

Let $\mathcal{D}_\mathrm{SFT} = \{(x_i, y_i)\}_{i=1}^{N_\mathrm{SFT}}$ denote the SFT dataset with instruction-response pairs and $\mathcal{D}_\mathrm{RL} = \{(x_j, r_j)\}_{j=1}^{N_\mathrm{RL}}$ denote the RL dataset with prompts and associated reward signals. We use $p_\theta$ to denote the model distribution parameterized by $\theta \in \mathbb{R}^d$, and $\eta > 0$ to denote the learning rate for gradient-based updates. Throughout, $\|\cdot\|$ denotes the Euclidean norm 
and $\langle \cdot, \cdot \rangle$ denotes the standard inner product.
The key notation used throughout this paper is in Table~\ref{tab:notation} in Appendix~\ref{app:notations}.

\paragraph{Supervised Fine-Tuning (SFT).}
\label{sec:sft-obj}
The first stage of the standard post-training pipeline minimizes the cross-entropy loss over instruction-response pairs:
\begin{equation}
\mathcal{L}_\mathrm{SFT}(\theta) = \mathbb{E}_{(x,y) \sim \mathcal{D}_\mathrm{SFT}} [-\log p_\theta(y \mid x)]
\end{equation}
This objective encourages the model to assign high probability to the target response $y$ conditioned on the instruction $x$. 
The set of SFT-optimal parameters is $\Theta^*_{\mathrm{SFT}} \coloneqq \{\theta : \mathcal{L}_{\mathrm{SFT}}(\theta) = \mathcal{L}^*_{\mathrm{SFT}}\}$. When a representative is needed, we write $\theta^*_{\mathrm{SFT}}\in \Theta^*_{\mathrm{SFT}}.$

\paragraph{Reinforcement Learning (RL).}
\label{sec:rl-obj}
The second stage optimizes a reward signal while constraining deviation from the SFT checkpoint via a KL penalty:
\begin{equation}\label{eq:L_rl}
\mathcal{L}_\mathrm{RL}(\theta) = \beta \cdot D_{KL}(p_\theta(\cdot \mid x) \parallel p_{\theta_\mathrm{SFT}}(\cdot \mid x)) - \mathbb{E}_{x \sim \mathcal{D}_\mathrm{RL}} [R(x, \text{sample}(p_\theta(\cdot \mid x)))]
\end{equation}
where $\beta > 0$ is the KL penalty coefficient, $R(\cdot)$ is the reward function, and $D_{KL}(p \parallel q) = \sum_z p(z) \log \frac{p(z)}{q(z)}$ denotes the KL divergence. The first term penalizes distributional drift from the SFT checkpoint, while the second (negative reward) term encourages high reward; \emph{minimizing} this loss trades off the two.
We denote the RL-optimal parameter by $\theta_\mathrm{RL}^* = \arg\min_\theta \mathcal{L}_\mathrm{RL}(\theta)$.

\paragraph{Polyak-Lojasiewicz (PL) Condition.}
\label{sec:pl}
A differentiable function $f: \mathbb{R}^d \to \mathbb{R}$ satisfies the PL condition with parameter $\mu > 0$ if:
\begin{equation}
\frac{1}{2} \|\nabla f(\theta)\|^2 \geq \mu (f(\theta) - f^*)
\end{equation}
for all $\theta$, where $f^* = \min_\theta f(\theta)$.
Strong convexity implies the PL condition, but the converse does not hold. The PL condition allows for non-convex geometries away from the minimum while maintaining local sharpness that drives convergence.
Over-parameterized deep networks have been empirically observed to satisfy PL-like conditions during training~\citep{zhang2016understanding}. Recent theoretical work establishes that locally verifiable PL conditions persist in finite-width deep networks~\citep{aich2025lplr}, particularly in the neural tangent regime~\citep{jacot2018ntk} where the network behaves like a linear model in a high-dimensional feature space. For transformer-based language models, the PL constant $\mu_\mathrm{SFT}$ can be interpreted as measuring the curvature of the loss landscape around the SFT optimum: flatter landscapes (smaller $\mu_\mathrm{SFT}$) permit larger parameter displacements before significant loss degradation, while sharper landscapes (larger $\mu_\mathrm{SFT}$) are more sensitive to perturbations.

\subsection{RL-SFT Non-Decoupling}\label{sec:rl-sft}

Let $\theta_\mathrm{SFT}^* \in \Theta_\mathrm{SFT}^*$ denote the SFT-optimal parameters. 
The subsequent RL phase minimizes the RL loss $\mathcal{L}_\mathrm{RL}(\theta) = \beta D_{KL}(p_\theta \| p_{\theta_\mathrm{SFT}}) - \mathbb{E}\mathcal{R}_\mathrm{RL}(\theta)$ (cf. Eq.~\eqref{eq:L_rl}), and we write
$g_\mathrm{SFT} \coloneqq \nabla\mathcal{L}_\mathrm{SFT}$ and $g_\mathrm{RL} := \nabla L_\mathrm{RL}$ throughout.

\textbf{Decoupling} would require $\mathcal{L}_\mathrm{SFT}(\theta_\mathrm{RL}) \approx \mathcal{L}_\mathrm{SFT}(\theta_\mathrm{SFT}^*)$, i.e., RL preserves SFT capabilities while improving reward. \textbf{Non-decoupling} is the failure of this property.

\begin{theorem}[SFT-then-RL Non-decoupling]
\label{thm:sft-rl}
Let $\Theta^*_{\mathrm{SFT}} \coloneqq \{\theta : \mathcal{L}_{\mathrm{SFT}}(\theta) = \mathcal{L}^*_{\mathrm{SFT}}\}$
denote the set of SFT-optimal parameters, where
$\mathcal{L}^*_{\mathrm{SFT}} = \min_{\theta} \mathcal{L}_{\mathrm{SFT}}(\theta)$, and let
$\mathrm{dist}(\theta, \Theta^*_{\mathrm{SFT}}) \coloneqq
\min_{\theta^* \in \Theta^*_{\mathrm{SFT}}} \|\theta - \theta^*\|_2.$
Suppose $\mathcal{L}_{\mathrm{SFT}}$ is continuously differentiable with a Lipschitz gradient and satisfies the PL condition with constant
$\mu_{\mathrm{SFT}} > 0$ on a convex region containing both the RL trajectory
and $\Theta^*_{\mathrm{SFT}}$.
Then, after RL
optimization starting from $\theta_0 \in \Theta^*_{\mathrm{SFT}}$, the resulting parameter $\theta_{\mathrm{RL}}$ satisfies
\begin{equation}
\label{eq:thm21}
\mathcal{L}_{\mathrm{SFT}}(\theta_{\mathrm{RL}}) - \mathcal{L}^*_{\mathrm{SFT}}
\;\ge\;
\frac{\mu_{\mathrm{SFT}}}{2}\,
\mathrm{dist}(\theta_{\mathrm{RL}}, \Theta^*_{\mathrm{SFT}})^2 .
\end{equation}
The loss gap $\mathcal{L}_{\mathrm{SFT}}(\theta_{\mathrm{RL}}) - \mathcal{L}^*_{\mathrm{SFT}}$
is strictly positive when
$\theta_{\mathrm{RL}} \notin \Theta^*_{\mathrm{SFT}}$.
\end{theorem}

If $\mathcal{L}_{\mathrm{SFT}}$ has a unique minimizer
$\theta^*_{\mathrm{SFT}}$ (e.g., under strong convexity or an
identifiability assumption), then $\Theta^*_{\mathrm{SFT}} =
\{\theta^*_{\mathrm{SFT}}\}$ and \eqref{eq:thm21} reduces to the
point-wise form
$\mathcal{L}_{\mathrm{SFT}}(\theta_{\mathrm{RL}}) -
\mathcal{L}_{\mathrm{SFT}}(\theta^*_{\mathrm{SFT}})
\ge \frac{\mu_{\mathrm{SFT}}}{2}
\|\delta\|_2^2$,
where $\delta := \theta_{\mathrm{RL}} - \theta^*_{\mathrm{SFT}}$
denotes the parameter displacement induced by RL.

The bound scales quadratically with parameter-space displacement $\|\delta\|$. 
While the gradient cosine and norm ratio are dimensionless (invariant to gradient rescaling), the absolute magnitude of SFT degradation depends on the PL constant $\mu_\text{SFT}$ and base loss, both of which vary with model size.

\begin{remark}[Bidirectional Non-Decoupling]
\label{rmk:rl-sft}
Let $\theta_\text{RL}^*$ be a local minimum of $\mathcal{L}_\text{RL}$. Since $\nabla^2 \mathcal{L}_\text{RL}(\theta_\text{RL}^*) \succeq 0$ by the second-order optimality condition, any displacement $\delta$ induced by subsequent SFT satisfies $\Delta \mathcal{L}_\text{RL} \geq 0$ within a neighborhood where the Hessian remains PSD, SFT cannot improve RL performance.

If $\mathcal{L}_\text{RL}$ additionally satisfies local quadratic growth with constant $\mu_\text{RL}$ near $\theta_\text{RL}^*$, then $\Delta \mathcal{L}_\text{RL} \geq \frac{\mu_\text{RL}}{2}\|\delta\|^2$. We verify this empirically via Hutchinson-based spectral checks
(Appendix~\ref{app:pl-estimation}) rather than claiming it as a theorem, as global PL for $\mathcal{L}_{\mathrm{RL}}$ is difficult to guarantee.
\end{remark}
The proofs are given in Appendix~\ref{app:proofs}. Theorem~\ref{thm:sft-rl} and Remark~\ref{rmk:rl-sft} establish a no-free-lunch property: whichever phase comes second degrades the first. The problem is sequential optimization itself, not the ordering.
This raises a practical question: \textit{how long should RL training last before the damage outweighs the benefit?}

To provide some structural insights into how SFT-RL coupling creates a fundamental trade-off where coupling damage scales quadratically while RL benefit scales at most linearly, we make the following early-stage linear approximation: RL loss improves \emph{approximately linearly during early-stage training} (e.g., steps 1--100). In reality, RL loss curves are highly non-linear due to diminishing returns, reward sparsity, and landscape sharpening.

\begin{assumption}[RL early-stage linear approximation]
$\mathcal{L}_\mathrm{RL}$ decreases by roughly constant $c_\mathrm{RL} > 0$ per step, i.e., \begin{equation}
\mathcal{L}_\mathrm{RL}(\theta_\mathrm{RL}) \approx \mathcal{L}_\mathrm{RL}(\theta_\mathrm{SFT}^*) - c_\mathrm{RL} T_\mathrm{RL}.
\end{equation}
\end{assumption}

\textbf{Why this is an optimistic assumption.} We argue by \emph{best-case
analysis}: In reality,
RL benefit scales sub-linearly due to diminishing returns. Even if RL maintained its early-stage improvement rate indefinitely
(an unrealistic best-case scenario), the quadratic-vs-linear trade-off structure
still predicts that coupling damage eventually dominates.

\begin{proposition}[Optimal RL Duration]
\label{prop:optimal-steps}
Let $\theta_\mathrm{SFT}^*\in \Theta_\mathrm{SFT}^*$ be the SFT optimum after $T_\mathrm{SFT}$ steps, with PL constant $\mu_\mathrm{SFT}$. Assume RL runs for $T_\mathrm{RL}$ steps with 
learning rate $\eta$ and mean RL update direction $\bar g_{\mathrm{RL}} := -\frac{1}{T_{\mathrm{RL}}}\sum_{t=1}^{T_{\mathrm{RL}}} g_{\mathrm{RL}}^{(t)}$, where $g_\mathrm{RL}^{(t)} = \nabla \mathcal{L}_\mathrm{RL}(\theta^{(t)})$ denotes the RL loss gradient evaluated at the $t$-th step (note the minus sign: $\bar g_{\mathrm{RL}}$ is the direction RL training actually moves, i.e., gradient descent on $\mathcal{L}_\mathrm{RL}$).
Since each RL step moves along the negative loss gradient, the cumulative RL update is
$\delta = -\eta\sum_{t=1}^{T_{\mathrm{RL}}} g_{\mathrm{RL}}^{(t)}
= \eta \bar g_{\mathrm{RL}} T_{\mathrm{RL}}$,
substituting $\|\delta\| = \eta\|\bar g_{\mathrm{RL}}\| T_{\mathrm{RL}}$
into Theorem~\ref{thm:sft-rl} and defining
$\kappa \triangleq \mu_{\mathrm{SFT}}\eta^2\|\bar g_{\mathrm{RL}}\|^2$
for compactness, the total composite loss $\mathcal{L}_{total} = \mathcal{L}_\mathrm{SFT}(\theta_\mathrm{RL}) + \lambda \mathcal{L}_\mathrm{RL}(\theta_\mathrm{RL})$ (with trade-off weight $\lambda > 0$) satisfies:
\begin{equation}\label{eq:l-tot}
\mathcal{L}_{total} \geq \underbrace{\mathcal{L}_\mathrm{SFT}^* + \lambda \mathcal{L}_\mathrm{RL}(\theta_\mathrm{SFT}^*)}_{C\ (\text{constant w.r.t.\ } T_\mathrm{RL})} + \frac{\kappa}{2} T_\mathrm{RL}^2 - \lambda c_\mathrm{RL} T_\mathrm{RL}
\end{equation}
where $c_\mathrm{RL} > 0$ is the per-step RL improvement rate and $C$ is independent of $T_\mathrm{RL}$. Since $C$ does not affect the minimizer, the optimal RL duration is:
\begin{equation}
\label{eq:optimal-trl}
T_\mathrm{RL}^* = \frac{\lambda c_\mathrm{RL}}{\kappa} = \frac{\lambda c_\mathrm{RL}}{\mu_\mathrm{SFT} \eta^2 \|\bar{g}_\mathrm{RL}\|^2}.
\end{equation}
\end{proposition}

\textbf{Interpretation.}  Flatter SFT landscapes (smaller $\mu_\mathrm{SFT}$) permit longer RL; larger learning rates or gradient norms prescribe shorter RL. 
We note that $T^*_{\mathrm{RL}}$ is the optimizer of the
bound in \eqref{eq:l-tot} and should be read as an illustrative estimate of the true optimum rather than its exact characterization.
The key insight is that $T^*_{\mathrm{RL}}$ is not a fixed
hyperparameter: for a given trade-off weight $\lambda$, it depends on the landscape geometry that varies across models, tasks, and training progress. Whether RL actually improves or harms SFT depends on the reward-SFT alignment $\alpha_{RS}$ (Proposition~\ref{prop:threshold}).

The early-stage approximation is intentionally optimistic. It intentionally overestimates
the benefit of RL because $c_{RL}$ decreases dramatically with training.
Consequently, our theoretical prediction of the steps is a loose upper bound.

$T_\mathrm{RL}^*$ moves during training as landscape geometry changes (Section~\ref{sec:landscape}). RL sharpens the landscape, increasing $\mu_\mathrm{SFT}$ and shortening the optimal window. Every RL step beyond $T_\mathrm{RL}^*$ is actively destructive: coupling damage $\frac{\kappa}{2} T_\mathrm{RL}^2$ grows quadratically while RL benefit scales at most sub-linearly.

We note two caveats on the displacement identity
$\|\delta\| = \eta\|\bar g_{\mathrm{RL}}\|T_{\mathrm{RL}}$: the identity
itself is exact, but the magnitude of $\|\bar g_{\mathrm{RL}}\|$ depends on
the correlation structure of successive RL gradients. Under persistent
gradient directions (full-batch deterministic updates),
$\|\bar g_{\mathrm{RL}}\| \sim \|g\|$ and the damage grows quadratically in
$T_{\mathrm{RL}}$. Under decorrelated (near-orthogonal) successive
gradients, $\|\bar g_{\mathrm{RL}}\| \sim \|g\|/\sqrt{T_{\mathrm{RL}}}$
(cf.\ Appendix~\ref{app:proofs:propto}), and the damage grows only linearly. Even in this regime, the optimal duration remains finite: the cumulative RL benefit is
concave (the per-step improvement $c_{\mathrm{RL}}$ decays with training),
so damage growing linearly still eventually dominates a sub-linear benefit.
The clean quadratic form of \eqref{eq:l-tot} corresponds to the
persistent-direction regime.

\subsection{When Does Non-Decoupling Emerge?}
\label{sec:task-hardness}

Theorem~\ref{thm:sft-rl} establishes non-decoupling at the SFT optimum ($\nabla \mathcal{L}_\mathrm{SFT} = 0$), where the bound is purely second-order. In practice, SFT may not fully converge, and a first-order term can partially offset the damage. This raises a quantitative question: \emph{under what conditions can RL actually improve SFT loss?}

\begin{proposition}[Non-decoupling Threshold]
\label{prop:threshold}
Let $\theta_\mathrm{SFT}$ be the parameter after SFT and $\delta = \eta T_\mathrm{RL} \bar{g}_\mathrm{RL}$ the RL displacement. Define the \emph{reward-SFT alignment} $\alpha_{RS} \coloneqq -\frac{\nabla \mathcal{L}_\mathrm{SFT}^\top \bar{g}_\mathrm{RL}}{\|\nabla \mathcal{L}_\mathrm{SFT}\| \cdot \|\bar{g}_\mathrm{RL}\|}$ and the \emph{effective curvature} $\sigma_{\text{eff}}^2 \coloneqq \bar{g}_\mathrm{RL}^\top H_\mathrm{SFT}\,\bar{g}_\mathrm{RL} / \|\bar{g}_\mathrm{RL}\|^2$, where $H_\mathrm{SFT}$ is the Hessian of $\mathcal{L}_\mathrm{SFT}$ along the displacement (evaluated at an intermediate point; cf. Appendix~\ref{app:B5}). Assuming $H_{\mathrm{SFT}}$ is PSD along the RL direction, then:
\begin{equation}
\label{eq:threshold-condition}
\Delta\mathcal{L}_\mathrm{SFT} < 0 \quad \Longleftrightarrow \quad \alpha_{RS} > \frac{\eta T_\mathrm{RL}\, \sigma_{\text{eff}}^2}{2\,\|\nabla \mathcal{L}_\mathrm{SFT}\|} \|\bar{g}_\mathrm{RL}\| \;=:\; \tau_{\text{crit}}
\end{equation}
\end{proposition}

\textbf{Remark on the RL reward.} $\alpha_{RS}$ is determined by the relationship between RL and SFT objectives. When RL uses a reward substantially different from SFT loss (e.g., human preferences), $\alpha_{RS}$ can be small or negative. When reward partially aligns with SFT (both favor correct outputs), improvement is possible. If reward $= -\mathcal{L}_\mathrm{SFT}$, RL reduces to continued SFT and non-decoupling trivially vanishes.

\textbf{Geometric implications.} SFT flattens the landscape ($\sigma_{\text{eff}}^2 \downarrow$), lowering $\tau_{\text{crit}}$ and making improvement likely. RL sharpens ($\sigma_{\text{eff}}^2 \uparrow$), raising $\tau_{\text{crit}}$ and eventually triggering non-decoupling. This implies that RL can improve SFT only until the curvature grows enough to flip the inequality.

\section{Relation to Modern Post-Training Methods}
\label{sec:related}

Recent advances in post-training methods, including Direct Preference Optimization (DPO)~\citep{rafailov2024direct}, Group Relative Policy Optimization (GRPO)~\citep{shao2024grpo}, Unified Fine-Tuning (UFT)~\citep{uft2024}, Kahneman-Tversky Optimization (KTO)~\citep{ethayarajh2024kto}, Odds Ratio Preference Optimization (ORPO)~\citep{hong2024orpo}, and pure RL, have yielded substantial improvements to individual stages of the pipeline. However, as we show, none of these existing approaches fully escape the non-decoupling problem.

\paragraph{Sequential methods (DPO, GRPO).}
While DPO eliminates the need for a separate reward model, it still operates as a sequential stage after SFT. Our Theorem~\ref{thm:sft-rl} applies regardless of the source of the reward signal, meaning DPO is equally subject to the non-decoupling bound. Similarly, GRPO and its variants improve credit assignment within the RL phase, but do not address the fundamental gradient misalignment between the SFT and RL objectives.

\paragraph{Unified methods (UFT, KTO, ORPO).}
Methods such as UFT, KTO, and ORPO replace the sequential pipeline with simultaneous optimization of both objectives, which represents a promising step forward. However, given the near-orthogonality of the gradients and the extreme norm imbalance between them, 
for any naive convex combination of the two objectives, the RL gradient dominates as soon as its mixing weight exceeds $1/(1+\|g_\mathrm{RL}\|/\|g_\mathrm{SFT}\|) \approx 0.05$ (see Section~\ref{sec:landscape}) \citep{NEURIPS2024_fa69e968}. Naive joint training therefore cannot escape the coupling problem; explicit gradient decomposition is required to resolve conflicting updates.

\textbf{Gradient surgery}~\cite{yu2020gradient} is a multi-task optimization technique that resolves conflicting gradient directions. Given two objectives with gradients $g_1$ and $g_2$, naive gradient combination $g = \alpha g_1 + (1-\alpha)g_2$ is dominated by the larger-norm component when gradients are misaligned. Gradient surgery projects out the conflicting component: if the gradients point in opposing directions (i.e., $\cos(g_1, g_2) < 0$), the method replaces $g_2$ with $g_2 - \frac{g_2^\top g_1}{\|g_1\|^2}g_1$, which removes the part of $g_2$ that would increase $\mathcal{L}_1$. In our setting, this means applying surgery at each training step to remove the component of $g_\mathrm{RL}$ that conflicts with $g_\mathrm{SFT}$, ensuring that RL updates never increase SFT loss. The challenge is that our observed norm ratio ($\|g_\mathrm{RL}\|/\|g_\mathrm{SFT}\| \approx 18.4$) means the projection leaves only a small fraction of $g_{RL}$ intact, raising the question of whether sufficient RL signal survives the surgery.

\section{Empirical Evidence}

\subsection{Validating Non-decoupling}

We conduct experiments on Qwen3-0.6B \citep{qwen3technicalreport} across four tasks spanning reasoning (MATH-500), instruction following (Alpaca), dialog (HH-RLHF), and binary classification (SST-2).

\paragraph{Experimental setup.}

We run SFT for 1000--4000 steps; RL for 500--2000 steps with KL penalty $\beta \in \{0.01, 0.05, 0.1\}$. RL uses task-based reward (1 for correct, 0 for incorrect) via REINFORCE with baseline. See Appendix~\ref{app:setup} for full hyperparameters.

\paragraph{Evaluation protocol.} SFT loss is measured as cross-entropy on the validation set at three checkpoints: (1) \emph{Base}: pretrained model before any fine-tuning; (2) \emph{Post-SFT}: after supervised fine-tuning; (3) \emph{Post-RL}: after RL optimization from the SFT checkpoint. The key metric is $\Delta\mathcal{L}_\mathrm{SFT} = \mathcal{L}_\mathrm{SFT}^{(post\text{-}RL)} - \mathcal{L}_\mathrm{SFT}^{(post\text{-}SFT)}$, where $\Delta > 0$ indicates RL degraded SFT capabilities (non-decoupling).

To test whether non-decoupling emerges, we conduct the above experiment on SST-2 (100 validation samples), a sentiment classification task requiring genuine semantic understanding. Results are shown in Table~\ref{tab:scaling-sst2}, which shows RL degradation in all 6 configurations. 

\begin{table}[t]
\centering
\caption{Training step sensitivity on SST-2 (100 validation samples, task-based reward). RL causes SFT loss degradation in all 6 configurations, confirming non-decoupling.}
\label{tab:scaling-sst2}
\begin{center}
\begin{small}
\begin{tabular}{cc|cccc|cc}
\toprule
SFT Steps & RL Steps & Base Loss & SFT Loss & Post-RL Loss & $\Delta$ Loss & SFT Acc & RL Acc \\
\midrule
\multicolumn{8}{c}{\textit{Part A: Varying SFT steps (RL=500)}} \\
1000 & 500 & 11.364 & 0.067 & 4.417 & $+4.350$ & 0.52 & 0.52 \\
2000 & 500 & 11.364 & 0.042 & 3.117 & $+3.075$ & 0.52 & 0.52 \\
4000 & 500 & 11.364 & 0.051 & 4.812 & $+4.761$ & 0.52 & 0.52 \\
\midrule
\multicolumn{8}{c}{\textit{Part B: Varying RL steps (SFT=2000)}} \\
2000 & 500 & 11.364 & 0.042 & 3.117 & $+3.075$ & 0.52 & 0.52 \\
2000 & 1000 & 11.364 & 0.034 & 2.526 & $+2.492$ & 0.52 & 0.51 \\
2000 & 2000 & 11.364 & 0.057 & 2.849 & $+2.793$ & 0.52 & 0.52 \\
\bottomrule
\end{tabular}
\end{small}
\end{center}
\end{table}

To test whether non-decoupling is task-specific, we replicate the SFT$\to$RL pipeline on three diverse datasets: MATH-500, Alpaca, and HH-RLHF. For each dataset, we first train Qwen3-0.6B via SFT on the dataset's prompts and references for 400 steps, then apply RL (REINFORCE with $\beta = 0.01$ KL penalty) for 200 steps optimizing the corresponding task reward. We report the cross-entropy loss on a held-out validation set at three checkpoints: base model (pre-training only), after SFT, and after RL in Table~\ref{tab:multitask}, where SFT loss is shown to increase after RL across all task types.

\begin{table}
\centering
\caption{Task generalization results ($\beta=0.01$). SFT loss increases after RL across all task types.}
\label{tab:multitask}
\begin{center}
\begin{small}
\begin{tabular}{lcccccc}
\toprule
Dataset & Base Loss & SFT Loss & SFT Acc & SFT$\to$RL Loss & Post-RL Acc &  $\Delta$ Loss \\
\midrule
MATH-500 & 5.603 & 5.248 & 15\% & 9.108 & 1\% & +3.860 \\
Alpaca & 7.384 & 4.858 & - & 6.137 & - & +1.279 \\
HH-RLHF & 7.737 & 6.703 & - & 8.021 & - & +1.318 \\
SST-2 & 11.364 & 0.058 & 52\% & 4.687 & 52\% & +4.629 \\
\bottomrule
\end{tabular}
\end{small}
\end{center}
\end{table}

\subsection{Loss Landscape Analysis}\label{sec:landscape}

To understand \emph{why} coupling in Theorem~\ref{thm:sft-rl} occurs in
practice, we analyze two geometric signatures: gradient misalignment
and the opposing flattening/re-sharpening dynamics of the SFT landscape across the two phases.

\begin{wrapfigure}{r}{0.65\textwidth}
\centering
\includegraphics[width=0.65\textwidth]{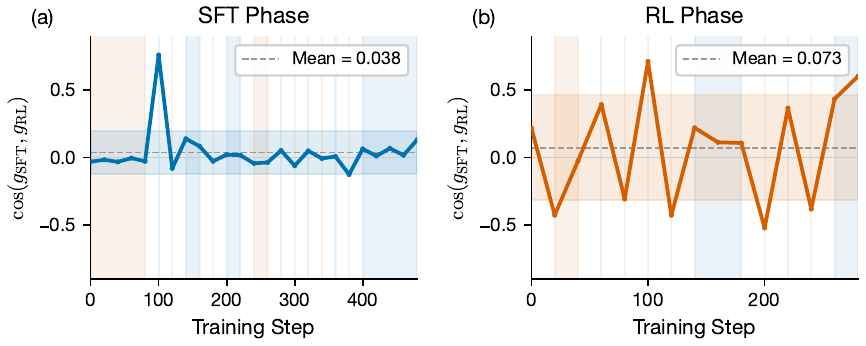}
\caption{Cosine similarity between SFT and RL gradients on MATH-500 (Qwen3-0.6B). \textbf{(a)} (SFT phase, first 500 steps): $\cos(g_\mathrm{SFT}, g_\mathrm{RL}) \approx 0.038 \pm 0.159$, indicating persistent near-orthogonality, i.e., SFT and RL optimize in different parameter subspaces. \textbf{(b)} (RL phase, steps 1--300): Cosine oscillates in $[-0.52, +0.71]$ (mean $0.073 \pm 0.389$) with $\|g_\mathrm{RL}\|/\|g_\mathrm{SFT}\| \approx 18.4$, meaning RL displaces parameters in large, nearly perpendicular steps---the worst case for SFT preservation (Theorem~\ref{thm:sft-rl}).}
\label{fig:gradient-dynamics}
\end{wrapfigure}

We first measure coupling via gradient alignment:
$
\cos(\phi) = \frac{\langle \nabla \mathcal{L}_\mathrm{SFT}, \nabla \mathcal{L}_\mathrm{RL} \rangle}{\|\nabla \mathcal{L}_\mathrm{SFT}\| \cdot \|\nabla \mathcal{L}_\mathrm{RL}\|}.
$
%
We record both gradients at each training step on MATH-500 (Qwen3-0.6B) and compute their cosine similarity. During SFT (1000 steps), we compute $\cos(g_\mathrm{SFT}, g_\mathrm{RL})$ by evaluating the RL gradient at each SFT checkpoint without applying it. After SFT converges, we switch to RL (300 steps, REINFORCE with $\beta = 0.01$) and compute $\cos(g_\mathrm{SFT}, g_\mathrm{RL})$ at each RL step using the frozen SFT loss gradient. We also track the gradient norm ratio $\|g_\mathrm{RL}\|/\|g_\mathrm{SFT}\|$ throughout both phases.
We observe persistent near-orthogonality (Figure~\ref{fig:gradient-dynamics}): $\cos(\phi) \approx 0.038$ during SFT, oscillating in $[-0.52, +0.71]$ during RL with an 18.4:1 gradient norm ratio ($\|g_\mathrm{RL}\|/\|g_\mathrm{SFT}\|$). RL updates are large and nearly perpendicular to SFT directions---precisely the worst case for Theorem~\ref{thm:sft-rl}.

Next, we examine the loss-landscape geometry through the SFT gradient
norm $\|\nabla \mathcal{L}_{\mathrm{SFT}}\|$, computed by
back-propagation. Along the SFT phase of the MATH-500 run, $\|g_{\mathrm{SFT}}\|$ declines from $95.6$ to $9.4$. We continue from the SFT checkpoint into 300 RL steps, and the norm rises from $15.5$ to
$60.9$, while the pointwise PL ratio $\hat\mu_{\mathrm{PL}}$ rises from $160$ to $2799$ (Appendix~\ref{app:pl-estimation}).

\textbf{Seesaw dynamics.} Table~\ref{tab:seesaw} reveals the geometric
mechanism: SFT flattens the landscape into a wide basin; subsequent RL
re-sharpens it. This geometric narrative explains Theorem~\ref{thm:sft-rl}: the lower bound on SFT loss is tight because the landscape geometry forces RL to climb steep walls in SFT-relevant directions: there is no gentle slope that allows RL to improve reward without harming SFT.

\section{Why We Need to Rethink the Sequential Paradigm}

The central claim of this position is that the SFT$\to$RL pipeline is suboptimal. The non-decoupling phenomenon is a mathematical consequence of optimizing misaligned objectives in sequence. We argue this on four grounds.

\begin{wraptable}{r}{0.5\textwidth}
\centering
\small
\caption{Seesaw dynamics of the SFT landscape (MATH-500, Qwen3-0.6B).
SFT flattens the landscape ($-90\%$ in the SFT gradient norm, 25
checkpoints); subsequent RL
re-sharpens it ($+293\%$). Norms computed
by back-propagation; the corresponding $\hat\mu_{\mathrm{PL}}$
trajectory is in Appendix~\ref{app:pl-estimation}.}
\label{tab:seesaw}
\begin{center}
\begin{tabular}{lccc}
\toprule
 & start & end & $\Delta$  \\
\midrule
SFT phase: $\|g_{\mathrm{SFT}}\|$ & $95.6$ & $9.4$ & $-90\%$ \\
RL phase: $\|g_{\mathrm{SFT}}\|$ & $15.5$ & $60.9$ & $+293\%$ \\
\bottomrule
\end{tabular}
\end{center}
\end{wraptable}

1. \textbf{The optimal stopping point is a moving target.}
Even if one could compute $T_\mathrm{RL}^*$ exactly at the start of RL training, it would be stale by the time it matters. RL itself reshapes the loss landscape (Section~\ref{sec:landscape}), increasing $\mu_\mathrm{SFT}$ and shortening the optimal window. This creates a paradox: the better RL works at improving reward, the faster it destroys the SFT foundation it depends on.

2. \textbf{The paradigm creates perverse incentives.}
Consider the strategic implication: if stronger SFT creates a sharper optimum that RL will subsequently degrade, practitioners are implicitly incentivized to train \emph{weaker} SFT models---ones that sit in flatter basins and thus absorb RL displacement with less damage. This is the opposite of what the community should want. The sequential pipeline punishes the very SFT quality it depends on, creating a destructive feedback loop between the two phases.

3. \textbf{Compute spent past $T_\mathrm{RL}^*$ is harmful.}
In standard training, stopping early saves compute but forgoes improvement. Post-training is worse: every RL step beyond the (unknowable) optimal point actively degrades model quality. Resources are wasted to make models worse.

\subsection{Alternative Paradigms}

1. \textbf{Joint optimization with coupling awareness.}
The natural fix is to optimize SFT and RL losses simultaneously: $(1\!-\!\alpha)\mathcal{L}_\mathrm{SFT} + \alpha\mathcal{L}_\mathrm{RL}$. But naive combination fails as our gradient analysis shows the objectives are nearly orthogonal, so simply adding gradients produces noise, not improvement. Gradient surgery~\cite{yu2020gradient} can remove conflicting components but requires per-step decomposition that is expensive at scale. A more promising direction is bilevel optimization~\cite{chen2025beyond}, where SFT provides a moving baseline that RL must improve without violating. Dynamic $\alpha$-scheduling, informed by Proposition~\ref{prop:optimal-steps}, could adaptively shift weight as landscape geometry changes.

2. \textbf{Landscape-aware scheduling.}
A less invasive fix: keep the sequential pipeline but monitor coupling in real time. Track $\cos(g_\mathrm{SFT}, g_\mathrm{RL})$ and estimate $\mu_\mathrm{SFT}$, computing $T_\mathrm{RL}^*$ online and stopping before damage accumulates. This is the ``early stopping'' approach grounded in our theoretical bound rather than validation loss heuristics. The limitation is that it treats the symptom (overshooting) rather than the cause (misaligned objectives).

3. \textbf{Eliminating the coupling entirely.}
Another alternative is to remove one of the two phases. DeepSeek-R1~\cite{deepseekr1} demonstrates that pure RL, without SFT, can produce capable models at 671B scale. The open question is the SFT-necessity threshold~\cite{qin2025supervisedfinetuningcurated}: below what model size or data scale does RL alone suffice? Conversely, reward-shaped SFT or constitutional AI~\cite{bai2022constitutional} could absorb RL's alignment function into a single phase, bypassing coupling entirely. The choice between these paths is itself an open empirical question.

\subsection{Open Questions}

We close with three questions that define the research agenda:
\begin{itemize}
\item[(Q1)] What are \emph{necessary and sufficient} conditions for decoupling? Our PL-condition bound is sufficient but likely not necessary.
\item[(Q2)] Can $\mu_\mathrm{SFT}$ and $\bar{g}_\mathrm{RL}$ be estimated online with enough precision to compute $T_\mathrm{RL}^*$ in practice?
\item[(Q3)] How should the community \emph{benchmark} coupling? We propose $\mathcal{C}(\theta) = \Delta\mathcal{L}_\mathrm{SFT}/\|\theta_\mathrm{RL} - \theta_\mathrm{SFT}^*\|^2$ as a standard metric, in the spirit of quantifying reward hacking via true-reward degradation \cite{skalse2022defining} and over-optimization scaling with optimization progress \cite{gao2023scaling}.
\end{itemize}
Regardless of which alternative prevails, three steps are actionable now: (1) joint optimization of the two objectives with gradient surgery to resolve conflicting updates; (2) landscape-aware scheduling that terminates RL training before damage accumulates; and (3) characterizing the model scale threshold above which SFT becomes unnecessary.

\section{Alternative Views}
\label{sec:objections}

1. \textbf{The PL condition is too strong.} PL is weaker than strong convexity and holds for over-parameterized networks~\cite{aich2025lplr,karimi2016}. Even without global PL, local quadratic growth near SFT minima suffices (Appendix~\ref{app:pl-validity}). Our qualitative predictions (gradient misalignment, curvature opposition) are \emph{empirical observations} that hold regardless of the PL constant $\mu$.

2. \textbf{$\cos \approx 0$ is a REINFORCE artifact.} The misalignment is between \emph{objectives}, not implementations. Any RL method optimizing a reward different from the SFT loss produces gradients in different eigensubspaces. Production PPO, GRPO, or DPO would exhibit similar coupling.

3. \textbf{DeepSeek-R1 shows that pure RL works.} DeepSeek-R1 demonstrates that pure RL without SFT can succeed at 671B scale, likely because Mixture-of-Experts models at that scale possess sufficient pretraining knowledge in terms of memorization
capacity to bypass SFT entirely. This counterexample suggests that a scale-dependent SFT-necessity threshold may exist. This is precisely our point: there may exist a threshold $\bar{N}$ above which SFT is unnecessary; capacity analyses of Transformers predict such model-data balance thresholds during pre-training \cite{niu2024beyond}. For models below $\bar{N}$, SFT provides essential knowledge distillation~\cite{qin2025supervisedfinetuningcurated}, and non-decoupling remains a first-order concern.

4. \textbf{The sequential paradigm may work at scale.} Our conclusions are constrained by the scope of our experiments. All results use Qwen3-0.6B, and validation at large scales is essential; the coupling coefficient $\kappa = \mu_\mathrm{SFT}\eta^2\|\bar{g}_\mathrm{RL}\|^2$ predicts that degradation grows with parameter count, but the precise scaling behavior remains empirical. The bound $\frac{\mu}{2}\|\delta\|^2$ applies for any $d$, predicting ${\sim}100\times$ larger coupling at 70B. 
Future investigations should measure $\mathcal{C}(\theta)$ and $\cos(g_\mathrm{SFT}, g_\mathrm{RL})$ during post-training on larger models (e.g., 70B).

5. \textbf{Production models work fine.} No major deployment logs $\Delta\mathcal{L}_\mathrm{SFT}$ at each RL checkpoint, and a low-cost diagnostic that would reveal coupling immediately. The same quality could likely be achieved at 30--50\% lower compute by stopping at $T_\mathrm{RL}^*$.

\section{Conclusion and Broader Impacts}

We have argued that the sequential SFT-then-RL paradigm suffers from a structural flaw that cannot be resolved by hyperparameter tuning. Under the PL condition, which holds for over-parameterized networks, RL degrades SFT performance by at least $\frac{\mu}{2}\mathrm{dist}(\theta_{\mathrm{RL}}, \Theta^*_{\mathrm{SFT}})^2$. The root cause is gradient misalignment: SFT and RL optimize in nearly orthogonal parameter subspaces ($\cos \approx 0$), so every RL step displaces parameters in a direction largely irrelevant to SFT preservation. The KL penalty $\beta$ constrains how far the model drifts in distribution space but cannot correct this directional mismatch in parameter space.

Our experiments on Qwen3-0.6B confirm the theory across four non-trivial tasks. The mechanism is measurable and consistent: 
SFT flattens the loss landscape into a wide basin while RL
re-sharpens it, and this seesaw dynamic manifests as a tightening PL
bound that converts early RL progress into irreversible SFT degradation.
The quadratic-versus-linear trade-off between coupling damage and RL benefit implies a finite optimal RL duration $T_{RL}^*$, but this optimum depends on landscape geometry that shifts during training itself, making any fixed schedule inherently suboptimal.

Three directions emerge from this analysis. First, joint optimization with explicit gradient decomposition can remove conflicting update components. Second, landscape-aware scheduling can monitor coupling in real time and stop RL before damage overwhelms benefits. Third, a SFT-necessity threshold may exist, and characterizing it would tell us when single-phase training suffices.

\paragraph{Broader Impacts.} Recognizing non-decoupling could reduce post-training compute by eliminating destructive RL steps, lowering both carbon emissions and barriers for smaller labs that cannot afford extensive hyperparameter searches. More broadly, understanding that sequential post-training has a structural flaw redirects research efforts from hyperparameter tuning toward paradigm-level innovation: joint optimization, landscape-aware scheduling, or single-phase training.

\section*{Acknowledgements and Disclosures}
The authors received no third-party funding for this work. The authors have no competing interests to disclose beyond their employment.

\bibliography{references.bib}
\bibliographystyle{plain}

\clearpage

\appendix

\section{Notations}
\label{app:notations}

\begin{table}[h]
\centering
\caption{Summary of key notation used throughout this paper. We use $p_\theta$ for the model distribution, $\mathcal{L}$ for loss objectives, $\mu$ for PL constants, and Greek letters for hyperparameters.}
\label{tab:notation}
\begin{tabular}{ll}
\toprule
\textbf{Symbol} & \textbf{Description} \\
\midrule
$\theta$ & Model parameters \\
$p_\theta$ & Model distribution parameterized by $\theta$ \\
$\eta$ & Learning rate \\
$\mathcal{D}_\mathrm{SFT}, \mathcal{D}_\mathrm{RL}$ & SFT and RL datasets \\
$\mathcal{L}_\mathrm{SFT}(\theta)$ & SFT cross-entropy loss \\
$\mathcal{L}_\mathrm{RL}(\theta)$ & RL loss (KL penalty $-$ reward) \\
$\beta$ & KL penalty coefficient in RL objective \\
$\Theta_\mathrm{SFT}^*$ & SFT-optimal parameter set \\
$\theta_\mathrm{SFT}^*$ & SFT-optimal parameter \\
$\theta_\mathrm{RL}^*$ & RL-optimal parameter \\
$D_{KL}(p \| q)$ & KL divergence $\sum_z p(z) \log \frac{p(z)}{q(z)}$ \\
$\mu_\mathrm{SFT}, \mu_\mathrm{RL}$ & PL constants for SFT and RL losses \\
$g_\mathrm{SFT}, g_\mathrm{RL}$ & Gradients of $\mathcal{L}_\mathrm{SFT}, \mathcal{L}_\mathrm{RL}$ \\
$\cos(g_\mathrm{SFT}, g_\mathrm{RL})$ & Gradient alignment (cosine similarity) \\
$H_\mathrm{SFT}, H_\mathrm{RL}$ & Hessians of SFT and RL losses \\
$\delta$ & Parameter displacement across a training phase, $\delta = \theta_\mathrm{RL} - \theta^*_\mathrm{SFT}$ (Section~\ref{sec:rl-sft})  \\
$\kappa$ & Coupling coefficient $\mu_\mathrm{SFT} \eta^2 \bar{g}^2$ \\
$T_\mathrm{SFT}, T_\mathrm{RL}$ & Number of SFT / RL training steps \\
$T_\mathrm{RL}^*$ & Theoretically optimal RL duration (Proposition~\ref{prop:optimal-steps}) \\
\bottomrule
\end{tabular}
\end{table}

\section{Proofs}
\label{app:proofs}

\subsection{Derivation of $\|\delta\| \propto \sqrt{d}$}\label{app:proofs:propto}
Consider a model with $d$ parameters, assuming that gradient variance does not decrease with $d$, which holds for standard architectures where the parameter count increases via wider layers (each receiving comparable gradient signals). The per-step gradient $g \in \mathbb{R}^d$ has entries with variance $\text{Var}(g_i) = O(1)$. Therefore, $\|g\|^2 = \sum_{i=1}^d g_i^2 \approx d \cdot \sigma_g^2$ where $\sigma_g^2$ is the average per-parameter variance. This gives $\|g\| \propto \sqrt{d}$. Over $T$ steps with learning rate $\eta$, the cumulative displacement $\delta = \eta \sum_{t=1}^T g_t$. In the high-dimensional regime ($d \gg 1$), successive gradients are approximately orthogonal so $\|\delta\|^2 \approx \eta^2 T \|g\|^2 \propto \eta^2 T d$, giving $\|\delta\| \propto \sqrt{d}$.
Consistent with the vector-average view of Section~\ref{sec:rl-sft}, when successive
gradients are decorrelated (near-orthogonal), the mean gradient satisfies
$\|\bar g\| \sim \|g\|/\sqrt{T}$, and the identity
$\|\delta\| = \eta\|\bar g\|T \sim \eta\sqrt{T}\|g\|$ recovers the $\sqrt{T}$ scaling derived above; when gradient directions persist across
steps, $\|\bar g\| \sim \|g\|$ and the displacement grows linearly in $T$,
as assumed in Appendix~\ref{sec:app:B4}.

\subsection{Proof of Theorem~\ref{thm:sft-rl} (SFT-then-RL Non-decoupling)}

The PL condition states that for all $\theta$,
\begin{equation}
\label{eq:pl-appendix}
\|\nabla \mathcal{L}_{\mathrm{SFT}}(\theta)\|_2^2 \;\ge\;
2\mu_{\mathrm{SFT}}\bigl(\mathcal{L}_{\mathrm{SFT}}(\theta) - \mathcal{L}^*_{\mathrm{SFT}}\bigr).
\end{equation}
Karimi et al.~\cite{karimi2016} show that, when the PL inequality holds on such a region, it implies quadratic growth towards the minimizer set: for all $\theta$,
\begin{equation}
\label{eq:qg-set}
\mathcal{L}_{\mathrm{SFT}}(\theta) - \mathcal{L}^*_{\mathrm{SFT}}
\;\ge\; \frac{\mu_{\mathrm{SFT}}}{2}\,
\mathrm{dist}\bigl(\theta, \Theta^*_{\mathrm{SFT}}\bigr)^2,
\qquad
\mathrm{dist}\bigl(\theta, \Theta^*_{\mathrm{SFT}}\bigr) =
\min_{\theta^* \in \Theta^*_{\mathrm{SFT}}} \|\theta - \theta^*\|_2 .
\end{equation}

Applying \eqref{eq:qg-set} at $\theta = \theta_{\mathrm{RL}}$ yields
\eqref{eq:thm21} immediately.

If $\theta_{\mathrm{RL}} \notin \Theta^*_{\mathrm{SFT}}$, then
$\mathrm{dist}(\theta_{\mathrm{RL}}, \Theta^*_{\mathrm{SFT}}) > 0$, so the
right-hand side of \eqref{eq:thm21} is strictly positive and
$\mathcal{L}_{\mathrm{SFT}}(\theta_{\mathrm{RL}}) > \mathcal{L}^*_{\mathrm{SFT}}$: any parameter
change that leaves the minimizer set strictly increases the SFT loss. Points
with equal loss belong to $\Theta^*_{\mathrm{SFT}}$ by definition, so they
incur zero distance and zero gap.

\qed

\subsection{Discussion of Remark~\ref{rmk:rl-sft} (Bidirectional Non-Decoupling)}

At a local minimum $\theta_\text{RL}^*$, the Hessian satisfies $\nabla^2 \mathcal{L}_\text{RL}(\theta_\text{RL}^*) \succeq 0$ (second-order necessary condition). For sufficiently small displacement $\delta$, the second-order Taylor expansion gives:
\begin{equation}
\mathcal{L}_\text{RL}(\theta_\text{RL}^* + \delta) = \mathcal{L}_\text{RL}(\theta_\text{RL}^*) + \frac{1}{2}\delta^\top \nabla^2 \mathcal{L}_\text{RL}(\theta_\text{RL}^* + \xi\delta) \delta \geq \mathcal{L}_\text{RL}(\theta_\text{RL}^*)
\end{equation}
since the Hessian remains PSD in a neighborhood of a non-degenerate minimum.

\subsection{Proof of Proposition~\ref{prop:optimal-steps} (Optimal RL Duration)}\label{sec:app:B4}

After $T_\mathrm{RL}$ steps of RL with learning rate $\eta$, the parameter displacement from $\theta_\mathrm{SFT}^*$ is approximately:

\begin{equation}
\delta = \theta_\mathrm{RL} - \theta_\mathrm{SFT}^* = -\eta
\sum_{t=1}^{T_\mathrm{RL}} g_\mathrm{RL}^{(t)} = \eta \bar{g}_\mathrm{RL} T_\mathrm{RL}, \quad
\bar{g}_\mathrm{RL} :=
-\frac{1}{T_\mathrm{RL}}
\sum_{t=1}^{T_\mathrm{RL}} g_\mathrm{RL}^{(t)},
\end{equation}
where the last equality is an identity by the definition of $\bar g_{\mathrm{RL}}$
(the vector average), so
$\|\delta\|^2 = \eta^2\|\bar g_{\mathrm{RL}}\|^2 T_{\mathrm{RL}}^2$.
Here we approximate
$\mathrm{dist}(\theta_{\mathrm{RL}}, \Theta^*_{\mathrm{SFT}})$ by the
cumulative displacement $\eta\bar \|g_{\mathrm{RL}}\|T_{\mathrm{RL}}$; this is
exact when the minimizer is unique (cf.\ the remark after Theorem~\ref{thm:sft-rl}) or
when the RL trajectory does not curve back toward the minimizer set.

From Theorem~\ref{thm:sft-rl}, the SFT loss increase is:

\begin{equation}
\Delta \mathcal{L}_\mathrm{SFT} = \mathcal{L}_\mathrm{SFT}(\theta_\mathrm{RL}) - \mathcal{L}_\mathrm{SFT}^* \geq \frac{\mu_\mathrm{SFT}}{2} \|\delta\|^2 = \frac{\mu_\mathrm{SFT}}{2} (\eta \bar{g}_\mathrm{RL} T_\mathrm{RL})^2 = \frac{\mu_\mathrm{SFT} \eta^2 \|\bar{g}_\mathrm{RL}\|^2}{2} T_\mathrm{RL}^2 = \frac{\kappa}{2} T_\mathrm{RL}^2
\end{equation}

where we define $\kappa = \mu_\mathrm{SFT} \eta^2 \|\bar{g}_\mathrm{RL}\|^2$ as the coupling coefficient.

For the RL loss, we introduce an abstract parameter $c_\mathrm{RL} > 0$ to denote the
\emph{per-step RL improvement rate}, capturing how much the RL loss decreases per training
step. 
\begin{equation}
\mathcal{L}_\mathrm{RL}(\theta_\mathrm{RL}) \approx \mathcal{L}_\mathrm{RL}(\theta_\mathrm{SFT}^*) - c_\mathrm{RL} T_\mathrm{RL}
\end{equation}

Let $f(T_{RL}) = \frac{\kappa}{2}T_{RL}^2 - \lambda c_\mathrm{RL}T_\mathrm{RL},$
we have
\begin{equation}
\frac{df}{dT_\mathrm{RL}} = \kappa T_\mathrm{RL} - \lambda c_\mathrm{RL} = 0 \implies T_\mathrm{RL}^* = \frac{\lambda c_\mathrm{RL}}{\kappa} = \frac{\lambda c_\mathrm{RL}}{\mu_\mathrm{SFT} \eta^2 \|\bar{g}_\mathrm{RL}\|^2}
\end{equation}
\qed

\subsection{Proof of Proposition~\ref{prop:threshold} (Non-decoupling Threshold)}\label{app:B5}

By second-order Taylor expansion of $\mathcal{L}_\mathrm{SFT}$ around $\theta_\mathrm{SFT}$:
\begin{equation}
\mathcal{L}_\mathrm{SFT}(\theta_\mathrm{RL}) = \mathcal{L}_\mathrm{SFT}(\theta_\mathrm{SFT}) + \nabla \mathcal{L}_\mathrm{SFT}^\top \delta + \frac{1}{2}\delta^\top H_{SFT} \delta + O(\|\delta\|^3)
\end{equation}
where $H_{SFT} = \nabla^2 \mathcal{L}_\mathrm{SFT}(\theta_\mathrm{SFT} + \xi\delta)$ for some $\xi \in (0,1)$.

\textbf{First-order term.} Substituting $\delta = \eta T_\mathrm{RL} \bar{g}_\mathrm{RL}$:
\begin{equation}
\nabla \mathcal{L}_\mathrm{SFT}^\top \delta = \eta T_\mathrm{RL} \nabla \mathcal{L}_\mathrm{SFT}^\top \bar{g}_\mathrm{RL} = -\eta T_\mathrm{RL} \alpha_{RS} \|\nabla \mathcal{L}_\mathrm{SFT}\| \cdot \|\bar{g}_\mathrm{RL}\|
\end{equation}
by definition of $\alpha_{RS}$.

\textbf{Second-order term.} The second-order term involves the Rayleigh quotient of the Hessian along the RL direction: $\sigma_{\text{eff}}^2 = \bar{g}_\mathrm{RL}^\top H_{SFT}\, \bar{g}_\mathrm{RL} / \|\bar{g}_\mathrm{RL}\|^2$.
\begin{equation}
\frac{1}{2}\delta^\top H_{SFT}\, \delta = \frac{\sigma_{\text{eff}}^2}{2}\|\delta\|^2 = \frac{\sigma_{\text{eff}}^2}{2}\eta^2 T_\mathrm{RL}^2 \|\bar{g}_\mathrm{RL}\|^2
\end{equation}

\textbf{Combining:}
\begin{equation}
\Delta\mathcal{L}_\mathrm{SFT} = -\eta T_\mathrm{RL}\, \alpha_{RS}\, \|\nabla \mathcal{L}_\mathrm{SFT}\| \cdot \|\bar{g}_\mathrm{RL}\| + \frac{\sigma_{\text{eff}}^2}{2}\eta^2 T_\mathrm{RL}^2 \|\bar{g}_\mathrm{RL}\|^2
\end{equation}

For improvement ($\Delta\mathcal{L}_\mathrm{SFT} < 0$), the negative first-order term must dominate over the positive second-order term:
\begin{equation}
\eta T_\mathrm{RL} \alpha_{RS} \|\nabla \mathcal{L}_\mathrm{SFT}\| \cdot \|\bar{g}_\mathrm{RL}\| > \frac{\sigma_{\text{eff}}^2}{2}\eta^2 T_\mathrm{RL}^2 \|\bar{g}_\mathrm{RL}\|^2
\end{equation}

Dividing both sides by $\eta T_\mathrm{RL} \|\nabla \mathcal{L}_\mathrm{SFT}\| \cdot \|\bar{g}_\mathrm{RL}\|$:
\begin{equation}
\alpha_{RS} > \frac{\sigma_{\text{eff}}^2 \,\eta T_\mathrm{RL} \|\bar{g}_\mathrm{RL}\|}{2 \|\nabla \mathcal{L}_\mathrm{SFT}\|} = \tau_{\text{crit}}
\end{equation}
\qed

\section{Experimental Setup Details}
\label{app:setup}

\textbf{Model.} We use Qwen3-0.6B, a 0.6B parameter transformer. This model size was chosen to enable systematic ablation studies (6+ training configurations per task) within feasible compute budgets, while being large enough to exhibit meaningful SFT-RL dynamics.

\textbf{Training hyperparameters (consistent across all experiments):}
\begin{itemize}
  \item \textbf{Optimizer:} AdamW with $\beta_1=0.9$, $\beta_2=0.999$, weight decay $=0.01$;
  \item \textbf{Learning rate:} $5 \times 10^{-5}$ (constant, no schedule);
  \item \textbf{Batch size:} 4 per step, gradient accumulation over 4 steps (effective batch 16);
  \item \textbf{KL penalty:} $\beta \in \{0.01, 0.05, 0.1, 1.0\}$ depending on the experiment;
  \item \textbf{RL objective:} Policy gradient (REINFORCE) with baseline, reward $=$ task accuracy (correct/incorrect prediction);
  \item \textbf{Precision:} Mixed precision (FP16/BF16) with gradient scaling.
\end{itemize}

\textbf{Datasets.}
\begin{itemize}
  \item \textbf{MATH-500:} 500 competition mathematics problems (100 sampled for training/evaluation). Tests mathematical reasoning.
  \item \textbf{Alpaca:} 52K instruction-following examples. Tests general instruction compliance. We evaluate on a held-out subset.
  \item \textbf{HH-RLHF:} Helpful and harmless dialog pairs from Anthropic. Tests conversational alignment.
  \item \textbf{COLA:} Corpus of Linguistic Acceptability from GLUE (10 validation samples). Simple binary classification for ablation baseline.
  \item \textbf{SST-2:} Stanford Sentiment Treebank (100 validation samples). Moderate-difficulty binary classification for ablation.
\end{itemize}

\section{When Does the PL Condition Hold?}
\label{app:pl-validity}

Our theoretical results rely on the Polyak-\L ojasiewicz (PL) condition. Here we discuss when this assumption is reasonable for neural network loss landscapes.

\textbf{Definition.} A function $f: \mathbb{R}^d \to \mathbb{R}$ satisfies the PL condition with constant $\mu > 0$ if:
\begin{equation}
\frac{1}{2}\|\nabla f(\theta)\|^2 \geq \mu (f(\theta) - f^*), \quad \forall \theta \in \mathbb{R}^d
\end{equation}
where $f^* = \inf_\theta f(\theta)$. This is weaker than strong convexity (which requires $\nabla^2 f \succeq \mu I$). Importantly, PL does \emph{not} require convexity: non-convex functions can satisfy PL locally.

\textbf{Why PL holds for SFT loss near convergence.} Several lines of evidence support the PL condition for cross-entropy loss near SFT optima:
\begin{enumerate}
  \item \textbf{Empirical landscape analysis:} Over-parameterized neural networks exhibit ``almost convex'' landscapes near minima~\cite{sagun2017empirical}. The Hessian spectrum has a bulk near zero and a small number of large outliers, with most directions exhibiting positive curvature. In our experiments, gradient norms decrease by $90\%$ during SFT
  (Table~\ref{tab:seesaw}), consistent with convergence to a basin where
  the gradient norm-to-loss gap ratio stabilizes (a hallmark of PL).
  \item \textbf{Over-parameterization and PL:} For sufficiently over-parameterized networks, the neural tangent kernel (NTK) regime~\cite{jacot2018ntk} guarantees PL-like conditions. Specifically, when the NTK $\Theta(\theta)$ satisfies $\Theta(\theta) \succeq \lambda_{\min} I$ with $\lambda_{\min} > 0$, the loss satisfies a PL condition with $\mu \geq \lambda_{\min}$~\cite{liu2022loss}. Our Qwen3-0.6B model, with 600M parameters trained on datasets of 100--52K examples, is in the over-parameterized regime where NTK analysis applies.
  \item \textbf{Local PL near minima:} Even if global PL fails, any smooth function with a non-degenerate minimum satisfies PL locally within a neighborhood where the Hessian is positive definite. Since SFT converges to a minimum with small gradient norm, the parameters reside in such a neighborhood, and Theorem~\ref{thm:sft-rl} applies for small displacements $\delta$.
\end{enumerate}

\textbf{When PL may fail.} The PL condition can fail in several regimes:
\begin{itemize}
  \item \textbf{Sharp minima with large flat regions:} If the SFT loss has a sharp minimum surrounded by extensive plateaus, $\mu$ may be large near the minimum but the QG bound may not hold far away. This could weaken our bounds for very long RL training (large $\|\delta\|$).
  \item \textbf{Non-differentiable objectives:} RL objectives with discrete rewards may not satisfy PL. Our theory applies to the smooth SFT loss landscape, not the RL objective.
  \item \textbf{Very early training:} Early in SFT training, the loss landscape may not satisfy PL. Our bounds apply after SFT has converged (which is the relevant regime for post-training).
\end{itemize}

\section{7B GRPO Bidirectional Validation}
\label{app:7b-grpo}

We have conducted experiments to test: (i) whether our
findings generalize beyond the 0.6B scale, (ii) whether they hold for algorithms other than REINFORCE, and (iii) what happens in the reverse (RL$\to$SFT) pipeline. We test both pipeline directions with GRPO on Qwen2.5-7B.

\subsection{Experimental Setup}
\label{app:7b-grpo-setup}

\begin{table}[h]
\centering
\label{tab:7b-grpo-setup}
\small
\begin{tabular}{ll}
\toprule
Component & Detail \\
\midrule
Model & Qwen2.5-7B \\
Task & MATH-500 (400 train / 100 test), 4-shot prompting \\
RL algorithm & GRPO: 50 steps, $K{=}8$ samples per prompt, $\beta{=}0.04$ (KL penalty), temp $=0.7$, lr $=5{\times}10^{-6}$ \\
SFT & 200 steps, lr $=2{\times}10^{-6}$ \\
Optimizer & 8-bit AdamW \\
Evaluation & Test loss (cross-entropy on MATH-500 test); accuracy (exact answer \\
 & match via \verb|\boxed{}| extraction, averaged over 50 problems) \\
\bottomrule
\end{tabular}
\end{table}

\subsection{Direction 1: SFT$\to$GRPO (Forward, Standard Pipeline)}
\label{app:7b-grpo-forward}

\begin{table}[h]
\centering
\caption{Forward pipeline (SFT then GRPO).}
\label{tab:7b-grpo-forward}
\small
\begin{tabular}{lcc}
\toprule
Stage & Test Loss & ACC \\
\midrule
Base (Qwen2.5-7B) & 0.817 & 36.0\% \\
After SFT & 0.546 & 44.0\% \\
After GRPO & 0.550 & 34.0\% \\
$\Delta$ (SFT$\to$GRPO) & $+0.004$ & $-10.0$ pts \\
\bottomrule
\end{tabular}
\end{table}

After GRPO following SFT, accuracy drops by 10 percentage points
($44\%\to 34\%$) while test loss slightly increases ($+0.004$): the
sequential SFT$\to$RL non-decoupling phenomenon is not specific to 0.6B or to
REINFORCE.

\subsection{Direction 2: GRPO$\to$SFT (Reverse Pipeline)}
\label{app:7b-grpo-reverse}

\begin{table}[h]
\centering
\caption{Reverse pipeline (GRPO then SFT).}
\label{tab:7b-grpo-reverse}
\small
\begin{tabular}{lcc}
\toprule
Stage & Test Loss & ACC \\
\midrule
Base (Qwen2.5-7B) & 0.817 & 36.0\% \\
After GRPO & 0.828 & 26.0\% \\
After SFT & 0.546 & 42.0\% \\
$\Delta$ (GRPO$\to$SFT) & $-0.282$ & $+16.0$ pts \\
\bottomrule
\end{tabular}
\end{table}

\textbf{Interpretation.} GRPO alone \emph{degrades} the base
model ($36\%\to 26\%$): we attribute this to Qwen2.5-7B's math ability
residing in a parameter-space region already shaped by an SFT-like objective;
with very limited effective GRPO training (see below), the reward-based policy
gradient perturbs the model away from this existing optimum rather than
improving on it. There were no GRPO accuracy gains in this direction, so the
subsequent SFT recovery to $42\%$ should not be described as
``overwriting GRPO gains''. Rather, SFT converges to essentially the same
optimum regardless of its starting point: $42\%$ closely matches the $44\%$
of SFT applied directly to the base model (Direction 1), i.e., the parameter
changes introduced by GRPO fall within SFT's basin of attraction and are
fully overwritten. In neither direction does the second objective retain the
first objective's learned behavior; this \emph{bidirectional overwrite} is
the core phenomenon our paper describes.

\subsection{Limitations}
\label{app:7b-grpo-limits}

The effective GRPO training here is far less than in our 0.6B experiments
(1000 REINFORCE steps): only 50 GRPO steps were run, and with group-relative
advantage estimation $\hat{A}_i = (r_i - \bar{r})/\sigma_r$, steps where all
$K{=}8$ samples receive identical rewards (all correct or all wrong) produce
zero gradient, resulting in a high skip rate. The small forward loss change
($+0.004$) is consistent with the small number of effective steps. Scaling up
GRPO training (more steps, higher temperature, larger $K$, importance
sampling) to produce stronger non-decoupling signals is an important future
direction that exceeds our current compute budget.

\section{Empirical PL Estimation}
\label{app:pl-estimation}
Theorem~\ref{thm:sft-rl} assumes that the SFT loss satisfies the PL condition near the SFT minimum. Appendix~\ref{app:pl-validity}
argues this qualitatively. Here we report empirical estimates
on the 0.6B model (Qwen3-0.6B, MATH-500 SFT): the PL ratio along SFT
(F.1), the Hessian spectrum (F.2), a quadratic-growth check (F.3),
and the evolution of the PL ratio during RL (F.4).

\subsection{Trajectory of the PL Constant During SFT}
We train SFT for 1000 steps, then continue for 2000 additional steps
to approximate the empirical lower bound of the training loss
($L^* \approx 0.2017$ at step 3000). At each SFT checkpoint, we
estimate the pointwise ratio
\[
\hat\mu_{\mathrm{PL}}(\theta_t) =
\frac{\|\nabla \mathcal{L}_{\mathrm{SFT}}(\theta_t)\|^2}
     {2\,(\mathcal{L}_{\mathrm{SFT}}(\theta_t) - \mathcal{L}^*)}.
\]

\begin{table}[h]
\centering
\small
\caption{Pointwise PL ratio $\hat\mu_{\mathrm{PL}}$ across SFT
checkpoints (Qwen3-0.6B, MATH-500). The ratio stays within a factor
of two of its mean ($182$) at every checkpoint.}
\begin{tabular}{lccccccc}
\toprule
SFT step & 50 & 100 & 200 & 300 & 500 & 750 & 1000 \\
\midrule
$\hat\mu_{\mathrm{PL}}$ & 243.5 & 187.2 & 207.1 & 159.2 & 132.2 &
178.9 & 164.3 \\
\bottomrule
\end{tabular}
\end{table}

Two caveats apply. First, pointwise positivity alone does not
establish a uniform PL constant: $\hat\mu_{\mathrm{PL}}(\theta_t) > 0$
holds at any point with $\nabla \mathcal{L} \neq 0$ and $\mathcal{L} > \mathcal{L}^*$, so the
meaningful signal is the \emph{stability} of the ratio: within a
factor of two ($[132, 244]$) with no downward drift. Second,
$\hat\mu_{\mathrm{PL}}$ inherits the approximation
$\mathcal{L}^* \approx \min_t \mathcal{L}_{\mathrm{SFT}}(\theta_t)$; a longer run would lower $\mathcal{L}^*$ and rescale all values. Together with the
quadratic-growth check below (F.3), the evidence is consistent
with a PL condition holding with a single constant along the SFT trajectory.

\subsection{Hutchinson Estimation of the Hessian Spectrum}
\label{app:hutchinson}

We estimate the Hessian spectrum of $\mathcal{L}_{\mathrm{SFT}}$ at SFT
checkpoints (steps 100, 500, 1000) with 100 random
Rademacher probes $\times$ 5 samples per probe and record the sign of the
Rayleigh quotient $g^{\top} \nabla^{2} \mathcal{L}_{\mathrm{SFT}}\, g$ for
each probe direction $g$.

\begin{table}[h]
\centering
\caption{Fraction of positive Rayleigh quotients from Hutchinson
estimation.}
\label{tab:hutchinson}
\small
\begin{tabular}{lccc}
\toprule
SFT step & 100 & 500 & 1000 \\
\midrule
Positive Rayleigh quotients & 99.0\% & 100\% & 97.4\% \\
\bottomrule
\end{tabular}
\end{table}

The results show that 97--100\% of Rayleigh quotients are positive, consistent with a positive semi-definite Hessian and hence with the local positive curvature that PL-based analyses presuppose near minima.

\subsection{Quadratic-Growth Verification}
\label{app:qg-check}
The PL condition implies quadratic growth towards the minimizer set
(Theorem~\ref{thm:sft-rl}), so we directly test this consequence:
using the SFT checkpoint at step 1000 as the reference $\theta^*$, we
compute
$
\frac{\mathcal{L}_{\mathrm{SFT}}(\theta_t) - \mathcal{L}_{\mathrm{SFT}}(\theta^*)}{\|\theta_t - \theta^*\|^2}
$
across the earlier SFT checkpoints (steps 50--750). If quadratic
growth holds with constant $c$, this ratio should be approximately
$c$ independent of $t$. The ratio is stable at
$4.0$--$4.6 \times 10^{-4}$ across all six checkpoints (coefficient
of variation $5.8\%$), consistent with the quadratic-growth
relationship. We note that the reference $\theta^*$ is a trajectory
checkpoint rather than a verified minimizer, so this is a consistency
check along the trajectory rather than a verification of global
quadratic growth.

\subsection{Evolution of the PL Ratio During RL}
\label{app:pl-rl}
We continue from a converged SFT checkpoint into 300 RL steps (REINFORCE, $\beta=0.1$) and track $\hat\mu_{\mathrm{PL}}(\theta_t)$ along the RL trajectory.

\begin{table}[h]
\centering
\small
\caption{SFT loss, gradient norm, and pointwise PL ratio during the
RL phase (Qwen3-0.6B, MATH-500). RL re-sharpens the SFT landscape:
$\|g_{\mathrm{SFT}}\|$ rises $3.9\times$ and $\hat\mu_{\mathrm{PL}}$
rises $17\times$ over 300 steps.}
\begin{tabular}{lccccccc}
\toprule
RL step & 0 & 50 & 100 & 150 & 200 & 250 & 300 \\
\midrule
$\|g_{\mathrm{SFT}}\|$ & 15.5 & 18.0 & 18.7 & 23.2 & 19.4 & 28.3 & 60.9 \\
$\hat\mu_{\mathrm{PL}}$ & 160 & 272 & 330 & 467 & 396 & 877 & 2799 \\
\bottomrule
\end{tabular}
\end{table}

The SFT-side PL constant grows by an order
of magnitude during RL: the bound of Theorem~\ref{thm:sft-rl}
tightens as RL proceeds, so the same parameter displacement costs
progressively more SFT loss.

\end{document}